# Invariant Cubature Kalman Filter for Monocular Visual Inertial Odometry with Line Features

**DELI YAN**[1,2,3]**, CHUNHUI WU**[2]**, WEIMING WANG**[2]**, YU SONG**[3]**, AND SHAOHUA LI**[1]

[1] State Key Laboratory of Mechanical Behavior and System Safety of Traffic Engineering Structures, Shijiazhuang Tiedao University, Shijiazhuang 050043, China
[2] School of Electrical and Electronic Engineering, Shijiazhuang Tiedao University, Shijiazhuang 050043，China
[3] School of Electronic and Information Engineering, Beijing Jiaotong University, Beijing 100044，China

Corresponding author: Deli YAN (e-mail: yandl@ stdu.edu.cn).

This work was supported in part by National Nature Science Foundation of China under Grant 11972238,Grant 11902206 and Grant 61573053,.in part by Natural Science Foundation of Hebei Province under Grant E2016210104，and in part by Science and Technology Research Project of Hebei Province under Grant Z2017022.

**ABSTRACT** To achieve robust and accurate state estimation for robot navigation, we propose a novel Visual Inertial Odometry(VIO) algorithm with line features upon the theory of invariant Kalman filtering and Cubature Kalman Filter (CKF). In contrast with conventional CKF, the state of the filter is constructed by a high dimensional Matrix Lie group and the uncertainty is represented using Lie algebra. To improve the robustness of system in challenging scenes, e.g. low-texture or illumination changing environments, line features are brought into the state variable. In the proposed algorithm, exponential mapping of Lie algebra is used to construct the cubature points and the re-projection errors of lines are built as observation function for updating the state. This method accurately describes the system uncertainty in rotation and reduces the linearization error of system, which extends traditional CKF from Euclidean space to manifold. It not only inherits the advantages of invariant filtering in consistency, but also avoids the complex Jacobian calculation of high-dimensional matrix. To demonstrate the effectiveness of the proposed algorithm, we compare it with the state-of-the-art filtering-based VIO algorithms on Euroc datasets. And  the results show that  the proposed algorithm is effective in improving accuracy and robustness of estimation.

## I. INTRODUCTION

Accurate and robust state estimation is of crucial importance for autonomous navigation. Visual inertial odometry is a technique to estimate the change of a mobile platform in position and orientation overtime using the measurement from on-board cameras and IMU sensor [1]. Visual information and inertial information have great complementarity in navigation system, which makes VIO widely used in Micro Aerial vehicles (MAV), self-driving cars and handheld devices.

Visual-inertial navigation system (VINS) has strong non-linearity, and the main source of non-linearity in the model is due to the kinematics of rotations. There are many ways to represent rotation, including Euler angles, rotation matrix, quaternion and Lie algebra methods. In these representations, Euler angle method is a minimal representation for the orientation, but it is not unique and has the problem of gimbal lock [2]. Rotation matrix uses a 3*3 matrix to represent rotation and have the problem of over-parametrization. Quaternion is widely used in VIO and SLAM areas in recent years, but it needs to be normalized to ensure its unit in each update step of filtering process [3]. More recently, Lie group representation for three-dimensional orientation/pose has become popular in SLAM, which can achieve better covariance and accuracy for both filter-based algorithm and optimization-based algorithms [4]. In this work, the state is expressed as a high dimensional matrix Lie group, and we mainly investigate how to transform the state and to estimate the posterior distribution of its mean and covariance.

Although many impressed results have been achieved in VIO area, the application of VIO still faces many challenging environments, such as low-textured, illumination changing scenes and some structured environment. In such cases, it is difficult to robustly detect and track points, but line information may be sufficient and can offer more for state estimation. So we investigate the possibility of estimating the 6-DOF pose a robot combing line feature and IMU data with a Kalman filter on Lie group.

In this paper, we provide a novel implementation of VIO estimation, named as L-SCKF-LG, which uses a high dimensional matrix Lie group including the line features to represent the state variable, and adopts square-root Cubature Kalman filter(SRCKF) to realize accurate estimation of robot pose. Different from traditional CKF methods, the proposed algorithm extends Cubature transformation from Euler space to Manifold space, and utilizes re-projection errors of line features to update the state, which maintains the advantage of consistence of invariant Kalman filter and improves the robustness in low-textured scenes. In summary, the main contributions of our work are:

- A novel VIO algorithm is proposed upon the theory of invariant Kalman filtering and Cubature Kalman filtering. The state in the filter is constructed by a high dimensional matrix Lie group and the uncertainty is represented using Lie algebra. Exponential mapping of Lie algebra is used to construct the cubature points, which extends traditional CKF from Euclidean space to manifold. The algorithm avoids complex computation of Jacobin Matrix and non-positive definite problem of covariance matrix in UKF
- The line features are included in the high dimensional matrix Lie group as part of the state, and the re-projection errors of lines are built as observation function for updating the state, which improve the robustness for pose estimation in low-textured environments.
- To demonstrate effectiveness of the proposed algorithm, we compare the proposed algorithm with state-of-the-art filtering-based VIO algorithms on Euroc datasets, and the detailed evaluation results are reported.

The remainder of this paper is organized as follows. In Section II, we give a survey of the related works. Section III mainly introduces some background knowledges relevant to our VIO approach. Section IV elucidates the detailed implementation of the proposed algorithm. Finally, experimental validation is presented in Section V, and conclusions are drawn in Section VI.

## II. RELATED WORKS

Over the last decades, significant processes have been achieved in visual-inertial odometry area. VIO system can be divided into different groups depend on various standards. According to coupling mode for visual and inertial information, VIO can be classified into two main streams: loosely-coupled and tightly-coupled approaches [8]. Tightly-coupled methods obtain more advantages in accuracy and robustness of estimation, so it gradually becomes the mainstream way of realization in recent years. An alternative classification approach is based on implementation mode, and VIO algorithms are divided into filtering-based approaches (e. g. MSCKF [9], ROVIO [10]) and optimization-based approaches, such as LSD-SLAM [11], OKVIS [12], VINS [13]. Optimization-based methods obtain the optimal estimate by jointly minimizing the residual using the measurements from both IMU data and images. Such approaches are able to achieve high accuracy at the cost of significant computational resources because of the iterative optimization process. In contrast, filter-based approaches generally use Kalman filter to predict and update the state, which are well suited to real time [5]. The link between filtering and optimization-based approaches can be built up within the framework of Bayesian inference[1]. In this work, the proposed algorithm adopts filtering-based and tightly-coupled method to estimate the state.

Recently, the Lie group and manifold representations for three-dimensional orientation have been widely used in VIO and SLAM areas, e.g., [14,15]. Both filter-based algorithms (e.g., [16,17]) and the optimization-based algorithms (e.g., [18,19]) can achieve better consistency and accuracy [20]. The inconsistency of conventional EKF-based VINS algorithms is mainly related to the partial observability, which means that these methods cannot observe the global translation and the rotation about the gravity direction [21]. The Invariant-EKF (IEKF) [22] was proposed by combining the symmetry-preserving theory and the extended Kalman filter, which made the traditional EKF possess the same invariance as the original system by using a geometrical adapted correction term and can effectively improve the consistency and accuracy. In [23], Right Invariant Extended Kalman Filter (RI-EKF) was proposed and applied into 2D SLAM. Zhang[6] applied RIEKF into 3D VINS owing to its consistency properties. In [24], RI-MSCKF was proposed, which was constructed by combining RIEKF and MSCKF algorithms. Brossard [5] built Right-UKF-LG algorithm upon Unscented Kalman Filtering on Lie group and IEKF, which spared the user the computation of Jacobians. But the UKF-computed covariance matrix is not always guaranteed to be positive define, which may cause the UKF to halt its operation [25]. The cubature approach is more accurate and more principled in mathematical terms than the sigma-point approach. So in this paper, we investigate the approach combining CKF and invariant Kalman filter to obtain more stable implementation.

Most of the exiting VIO approaches based on invariant Kalman filter only use the point features to update the state. In texture less or illumination changing environments, point tracking may face great challenging. Line features can provide more structural information on the environment than points [26]. Many impressive results have been achieved in recent years. Kottas and Roumeliotis [27] investigated the observability of the VIO using line features only. Kong [28] built a stereo VIO system combining point and line features by utilizing trifocal geometry. He [29] proposed PL-VIO algorithm, which optimized the state by minimizing a cost function combing IMU error term tighter with point and line re-projection error term. To my best knowledge, few literatures combine line features within invariant Kalman filter. To improve the robustness for invariant VIO algorithms in texture-less scene, we propose the L-SCKF-LG algorithms, which combine the line features into matrix Lie group as the state, and transform its uncertainty using SRCKF. The proposed algorithm achieves great performance for the state estimation.

## III. SOME PRELIMINARIES

This section mainly introduces some background knowledges which the proposed algorithm is concerned with, including the VIO coordinate frame, IMU kinematic model, the representation and measurement of 3D line, and the uncertainty representation for matrix Lie group.

### A. VIO COORDINATE FRAME

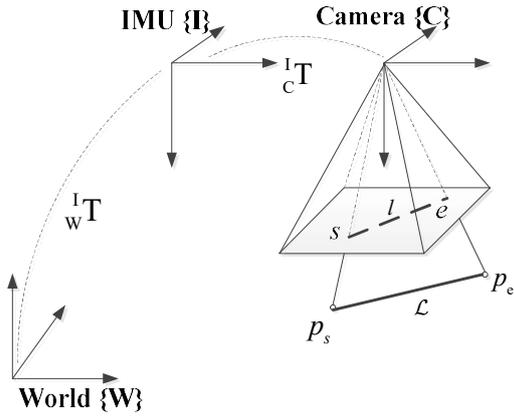

**FIGURE 1.** Coordinate frame for VIO

Three different coordinate frames are used throughout the paper: the world coordinate frame $\{W\}$, the IMU coordinate frame $\{I\}$ and the camera coordinate frame $\{C\}$. Figure 1 illustrates the spatial relations between different frames. In this figure, we denote symbol $^W_I T \in SE(3)$ as the transmission from inertial frame $\{I\}$ to world frame $\{W\}$, which includes the corresponding rotation matrix $^W_I R \in SO(3)$ and the position vector $p_I \in \mathbb{R}^3$. The symbol $^I_C T$ is the transformation between camera frame $\{C\}$ and inertial frame $\{I\}$, which is known in the provided datasets or from prior calibration. $\mathcal{L}$ is the spatial line with endpoints $(p_s, p_e)$, and $l$ is the projecting line in image. The problem we face is that of estimating the pose $^W_I T$ at any time using filtering method to merge visual and IMU information.

### B. IMU KINEMATIC MODEL

IMU can measure 3-axis acceleration and 3-axis angular velocity of the body at time $t$ with respect to the inertial frame $\{I\}$, which is notated as $^I\hat{a}(t)$ and $^I\hat{\omega}(t)$. The measurements are affected by additive white noise $n = [n_\omega, n_a] \in \mathbb{R}^6$ and sensor bias $b = [b_\omega, b_a] \in \mathbb{R}^6$ [14]:

$$^I\hat{\omega}(t) = {^I\omega(t)} + b_\omega(t) + n_\omega, \quad (1)$$

$$^I\hat{a}(t) = {^I_W R(t)}(^W a(t) - {^W g}) + b_a(t) + n_a. \quad (2)$$

where $^W g$ is the gravity vector in world frame $\{W\}$. The IMU state at time $t$ can be described by orientation $^W_I R \in SO(3)$, position $p_I \in \mathbb{R}^3$, velocity $^W V_I$ and bias $b$. The measured effects of the earth's rotation are small compared to the gyroscope accuracy, so the time evolution of the IMU state is similar to [30]:

$$^W_I \dot{R} = {^W_I R} \cdot (^I\omega)^\wedge, \quad ^W \dot{V}_I = {^W a}, \quad ^W \dot{p}_I = {^W V_I}. \quad (3)$$
$$\dot{b}_\omega = n_\omega, \quad \dot{b}_a = n_a$$

where the operator $(\cdot)^\wedge$ is a mapping from a vector in $\mathbb{R}^3$ to a skew symmetric matrix in $\mathbb{R}^{3\times 3}$.

In practice, the IMU samples the signals $^I\hat{a}$ and $^I\hat{\omega}$ with a fixed-frequency, and these measurements are used for state propagation in the filer. The time interval between two measurements is $\Delta t$. Assuming that the measurements $^I\hat{a}$ and $^I\hat{\omega}$ keep constant in $[t, t+\Delta t]$, the IMU propagation is calculated by:

$$\begin{cases} ^W_I R(t+\Delta t) = {^W_I R(t)} \cdot \exp(^I\omega(t)\Delta t) \\ ^W V(t+\Delta t) = {^W V(t)} + {^W a(t)}\Delta t \\ ^W p(t+\Delta t) = {^W p(t)} + {^W V(t)}\Delta t + \frac{1}{2} {^W a(t)}\Delta t^2 \end{cases} \quad (4)$$

This numerical calculation assumes a constant $^I\hat{a}$ and $^I\hat{\omega}$ between two measurements, which will generate integral calculation error. In practice, the use of a high-rate IMU mitigates the effects of this approximation[30], so the integration scheme we adopt is feasible.

### C. REPRESENTATION AND MEASUREMENT OF 3D LINE

(1) Representation of 3D line

Lines have 4 degrees of freedom in 3D space, which have many mathematical forms to represent. In this paper, we adopt two parameters for a 3D line. The line described by two endpoints is used in the filter, and Plücker coordinate is used for transformation and projection. The Plücker line can be defined from two points of the line as in [31].

The spatial points of the line can be denoted by $p = [x, y, z]^T$, whose homogeneous coordinate is $\bar{p} = [x, y, z, w]^T$. In Plücker coordinate, 3D spatial line can be represented by $\mathcal{L} = [\mathbf{n}^T, \mathbf{v}^T]^T \in \mathbb{R}^6$, where $\mathbf{v} \in \mathbb{R}^3$ is the line direction vector, and $\mathbf{n} \in \mathbb{R}^3$ is the normal vector of the plane determined by the line and the coordinate origin. The Plücker line can be constructed from two points on the line as follows:

$$\mathcal{L}=\begin{bmatrix}\mathbf{n}\\\mathbf{v}\end{bmatrix}=\begin{bmatrix}p_1\times p_2\\w_1 p_2 - w_2 p_1\end{bmatrix}\in\mathbb{R}^6, \quad (5)$$

where the Plücker constraint is $\mathbf{n}^T\mathbf{v}=0$.

(2) Frame transformation and pin-hole projection

The 3D line is projected to the image plane through pin-hole camera model, and the first step is to transform the line ${}^W\mathcal{L}$ represented in world frame $\{W\}$ to the camera frame $\{C\}$, denoted by ${}^C\mathcal{L}$. We assume that the transformation matrix is ${}^C_W T=\{{}^C_W R,\,{}^C_W p\}$, where ${}^C_W R$ is the rotation and ${}^C_W p$ is the transition between two frames. The transformation is shown in Equation(6) as described in[32],

$$^C\mathcal{L}=\begin{bmatrix}{}^c\mathbf{n}\\{}^c\mathbf{v}\end{bmatrix} = {}^C_W\mathcal{T}\cdot{}^W\mathcal{L}=\begin{bmatrix}{}^C_W R & \left({}^C_W p\right)_\times{}^C_W R\\0 & {}^C_W R\end{bmatrix}\cdot{}^W\mathcal{L}. \quad (6)$$

The projection process of Plücker line can be expressed in Equation (7),

$$^c l=\mathcal{K}\cdot{}^c\mathbf{n}=\begin{bmatrix}f_v & 0 & 0\\0 & f_u & 0\\-f_v c_u & -f_u c_v & f_u f_v\end{bmatrix}\cdot{}^c\mathbf{n}, \quad (7)$$

where $\mathcal{K}$ is the projection matrix for Plücker line, and the parameters in it can be got from camera intrinsic matrix. From the projection equation, we can find that only the normal vector ${}^c\mathbf{n}$ is used, and the direction vector ${}^C\mathbf{v}$ is omitted.

(3) Estimation of 3D line

The spatial line can be calculated from a pair of corresponding lines in two views, which is similar to Hartley and Zisserman's method[33]. We illustrate the estimation process in Figure 2.

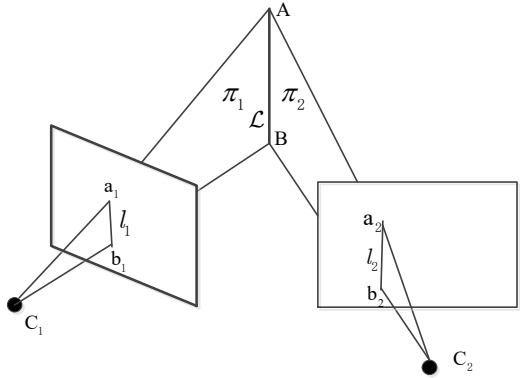

**FIGURE 2. Estimation of spatial line from a pair of corresponding lines in two views**

As shown in Figure 2, $l_1$ and $l_2$ are the projection of 3D line $\mathcal{L}$ on camera $C_1$ and $C_2$ respectively, which can be represented by end points $(a_1, b_1)$ and $(a_2, b_2)$. The 3D line is represented by two endpoints A and B, whose normalized coordinates $\bar{A}$ and $\bar{B}$ can be calculated using pinhole projection model. We can determine the plan $\pi_1$ through three non-collinear points, including $C_1$, $\bar{A}$ and $\bar{B}$, the process follows Equation (8),

$$\pi_1=\begin{pmatrix}(\bar{A}-C_1)\times(\bar{B}-C_1)\\-C_1^T(\bar{A}\times\bar{B})\end{pmatrix}. \quad (8)$$

The plan $\pi_2$ can also be obtained using the same way. Then the dual Plücker matrix $L^*$ can be computed by Equation (9),

$$L^* = \pi_1\pi_2^T - \pi_2\pi_1^T. \quad (9)$$

Then the Plücker line $\mathcal{L}=[\mathbf{n}^T,\mathbf{v}^T]^T$ can be extracted from the dual Plücker matrix $L^*$ following (10).

$$L^* = \begin{bmatrix}(\mathbf{v})^\wedge & \mathbf{n}\\-\mathbf{n}^T & 0\end{bmatrix}. \quad (10)$$

(4) Matrix lie group and uncertainty representation

In invariant Kalman filtering, the state space is represented by a matrix Lie group. A matrix Lie group $G\subset\mathbb{R}^{N\times N}$ is a subset of square invertible matrices with following properties:

$$I_N\in G;\ \forall\chi\in G,\chi^{-1}\in G;\ \forall\chi_1,\chi_2\in G,\chi_1\chi_2\in G, \quad (11)$$

where $I_N$ is the identity matrix. Lie algebra $\mathfrak{g}$ elements represent a tangent space of a group $G$ at the identity element[34]. The link between a Lie group and its associated Lie algebra can be expressed with the exponential (exp) and logarithm (log) operations. Rotation matrix $R\in SO(3)$ can be expressed by a vector $\xi\in\mathbb{R}^3$ through exponential and logarithm operations:

$$\begin{cases}R=\exp(\xi)=\exp_m(\xi^\wedge)\\\xi=\log(\exp(\xi))\end{cases}, \quad (12)$$

where $\exp_m(g)$ is exponential operation for matrix.

The traditional add operation is not close in Lie group, so the definition for random variable on Lie group described in [35] is adopted. The probability distribution of random variable $\chi\in G$ is expressed as $\chi\sim\mathcal{N}_r(\bar{\chi},P)$, with

$$\chi=\exp(\xi)\bar{\chi},\ \xi\sim\mathcal{N}(0,P), \quad (13)$$

where $\mathcal{N}(\cdot,\cdot)$ is the classical Gaussian distribution in Euclidean space, and $\mathcal{N}_r(\cdot,\cdot)$ is not Gaussian, which only represent "right" multiplication method. $\bar{\chi}$ is viewed as the mean, and $P$ is the covariance of the small noisy perturbation $\xi$ which describes the uncertainty of the variable.

## IV. THE PROPOSED VIO ALGORITHM

The VIO algorithm based on Kalman filter focus on estimating the current robot pose and the positions of all observed landmarks with the given motion model and the observation model. In this section, we mainly introduce the proposed algorithm with two main models: the propagation step and the upstate step. The propagation step predicts state and covariance using IMU measurement, and the upstate step estimates the posterior distribution of state and covariance using measurement of camera. Different from traditional filter-based algorithms, the state variable is built by a high dimensional Matrix Lie group including the elements to be estimated, and the corresponding uncertainty is propagated using square-root CKF implemented on manifold. It not only inherits the advantages of invariant filtering in consistency, but also avoids the complex Jacobian derivation process of high-dimensional matrix, which effectively improves the accuracy and robustness for VIO estimation in complex environment. In order to improve the performance of VIO running in low-texture environments, the line features are used to update the state. We present the detailed derivation of the proposed algorithm in subsections.

### A. THE STRUCTURE OF STATE VARIABLE

The structure of state variable is similar as in [5], and the difference is that the landmarks in the state are lines instead of 3D points. The variables we need to estimate are all included in the state, such as the orientation $R$, the position $p_I$, velocity $V$, IMU bias $b = [b_\omega, b_a]$ and 3D line features $p_l$. The variables, including $R$, $p_I$, $V$ and $p_l$, are built into a high dimensional matrix Lie group, defined as $\chi \in SE_{2+2m}(3)$, where $m$ is the number of line landmarks in the state. The structure of $\chi$ can be represented by

$$\chi = \begin{bmatrix} R & V & p_I & p_l^1 & \cdots & p_l^m \\ 0_{(2+2m)\times 3} & & I_{(2m+2)\times(2m+2)} & & \end{bmatrix}. \quad (14)$$

The 3D line in state is expressed by two end points, $p_l = [p_s, p_e]$, so the corresponding uncertainties for $\chi$ is defined as $\xi$, whose concrete structure is expressed as

$$\xi = [\xi_R^T \ \xi_V^T \ \xi_{p_I}^T \ \xi_{p_s^1}^T \ \xi_{p_e^1}^T \cdots \xi_{p_s^m}^T \ \xi_{p_e^m}^T]^T \in \mathbb{R}^{6m+9}. \quad (15)$$

Based on (13), $\chi$ can be represented by

$$\chi = \exp(\xi)\bar{\chi}, \quad \xi \sim \mathcal{N}(0, P_\chi), \quad (16)$$

where $P_\chi$ is the covariance of $\chi$.

The state also include the IMU bias $b$, which can be expressed by $b = \bar{b} + \tilde{b}$. $\bar{b}$ is the mean, and $\tilde{b}$ represent the uncertainty of $b$. So the whole state of filter can be represented by $X = [\chi, b]$, and the corresponding mean and uncertainty can be expressed as $\bar{X} = [\bar{\chi}, \bar{b}]$, $\sigma = [\xi, \tilde{b}]$.

At time $k$, the state $X_k$ following Gaussians distribution, and the covariance matrix is $P_k$. The structure of $P_k$ is expressed in Figure 3. $P_{p_{si}}$ and $P_{p_{ei}}$ represent positon covariance of two end points ($p_s$, $p_e$) for the $i$th line, the corresponding uncertainty vector are $\xi_{p_{si}}$ and $\xi_{p_{ei}}$.

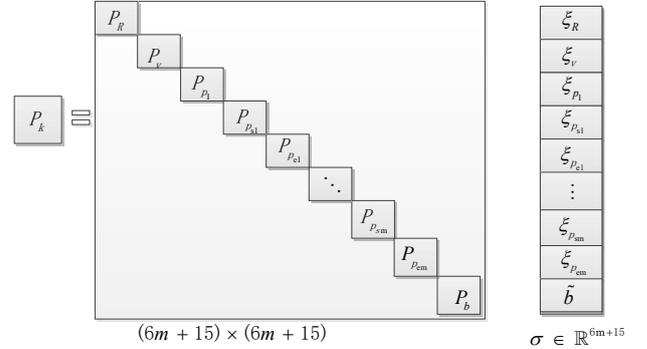

**FIGURE 3.** Covariance structure of state variables, and the corresponding uncertainty including the IMU bias.

### B. PROPAGATION STEP

The filter propagates the mean and covariance of the state using IMU kinematic model describing in section III.B. In our approach, the uncertainty was defined on Lie algebra, so we use the Cubature Kalman filter on Lie group to pass uncertainty through the compound pose change. Comparing with I-EKF, this method can avoid complex computation of Jacobian in the process of linearization.

(1) Construct the cubature points

Using the CKF methods to propagate the state, the first step is to compute the cubature points. Because the state $X$ in proposed algorithm is partially embedded into a Lie group, the process of computing cubature points is different from traditional CKF method[26].

At time $k$, the system's input is acceleration and angular velocity of body measured by IMU, denoted as $u_k = [\omega_k, a_k]$. The corresponding noise covariance is $Q_u$, whose square-root factor is denoted as $S_u$. To evaluate the cubature points, the state matrix and associated square-root factor are augmented with the $u_k$ and $S_u$, given by

$$X_{aug}^k = [X_k, u_k]^T \quad S_{aug}^k = \begin{bmatrix} S_k & \\ & S_u \end{bmatrix}. \quad (17)$$

The dimension of $S_{aug}$ is $l_s = 6m+21$, so the generator points $\{\zeta^j\}$ can be calculated by

$$\{\zeta^j\} = \sqrt{l_s} S_{aug}[1]_j, \quad j = 1 \ldots 2l_s, \quad (18)$$

where each $\zeta^j$ can be decomposed to the state uncertainty $\xi_k^j$, the bias uncertainty $\tilde{b}_k^j$ and input noise $n_u^j$, denoted as $\zeta^j = [\xi_k^j, \tilde{b}_k^j, n_u^j]^T$. Based on the spherical-radial rule described in [26], the state cubature point can be calculated by (19). We define the operator '$\oplus$', which include three parts. The IMU bias and noise can use the add operation in Euclidean space, and exponential mapping is used for $\xi_k^j$. Here the structure of cubature points in the proposed algorithm is different from traditional algorithms.

$$\overline{X}_{aug,k} \oplus \zeta^j \triangleq \begin{cases} \exp(\xi_k^j)\overline{\chi}_k \\ \overline{b}_k + \tilde{b}_k^j \\ u_k + n_u^j \end{cases}, j = 1 \ldots 2I_s. \quad (19)$$

(2) Propagate the cubature points and mean

The state cubature points pass through the IMU kinematic model described in (12), we use $f(\cdot)$ to represent the model. The transformed cubature points $[\chi_{k+1|k}^j, b_{k+1|k}^j]$ can be calculated by (20).

$$[\chi_{k+1|k}^j, b_{k+1|k}^j] = f(\exp(\xi_k^j)\overline{\chi}_k, u_k - b_k^j, n_u^j). \quad (20)$$

The state mean and bias mean also pass through the kinematic model $f(\cdot)$, and the predicted mean can be calculated by (21).

$$[\overline{\chi}_{k+1|k}, \overline{b}_{k+1|k}] = f(\overline{\chi}_k, u_k - \overline{b}_k, 0). \quad (21)$$

(3) Estimate the square-root factor of the predicted error covariance

The square-root Cubature Kalman filter is applied to estimate the state, so the predicted covariance square-root factor $S_{k+1|k}$ is essential. To compute $S_{k+1|k}$, a state and bias uncertainty vector $[\xi_{k+1|k}, \tilde{b}_{k+1|k}]^T$ is calculated through (22) and (23).

$$\xi_{k+1|k}^j = \frac{1}{\sqrt{2I_s}} \log(\chi_{k+1|k}^j \overline{\chi}_{k+1|k}^{-1}), j = 1 \ldots 2I_s \quad (22)$$

$$\tilde{b}_{k+1|k}^j = \frac{1}{\sqrt{2I_s}} (b_{k+1|k}^j - \overline{b}_{k+1|k}), j = 1 \ldots 2I_s \quad (23)$$

In (22), $\xi_{k+1|k}^j$ is the state uncertainty vector in $j$th cubature point, which is defined in Lie algebra, so it is computed using logarithm mapping. By performing QR decomposition on the matrix composed of state and IMU bias uncertainty, the $S_{k+1|k}$ can be calculated as the transpose of upper triangular matrix, that is

$$[q, r] = qr([\xi_{k+1|k}^1 \ldots \xi_{k+1|k}^{2I_s}; \tilde{b}_{k+1|k}^1 \ldots \tilde{b}_{k+1|k}^{2I_s}; 0 \quad S_u]) \quad (24)$$

$$S_{k+1|k} = r^T \quad (25)$$

## C. UPDATE STEP

In the proposed algorithm, the measurement of line feature is used for updating the state and covariance. In this section, we mainly introduce the measurement model for line features and the update process using cubature transformation on Lie group.

(1) Measurement model for line features

In this paper, the LSD detector[36] is used to detect the line segments in the new frame, and LBD descriptor[37] is adopted to match the lines between two images. The parameterization and the projection for 3D line are described in section III.(C). In the filter, the 3D line is described by two endpoints $(p_s, p_e)$, which is denoted as $^W\mathcal{L}$ with respect to the world frame $\{W\}$. At time $k+1$, the matched line segment in image is denoted as $s = (s_s, s_e)$, and the predicted line $l = [l_1, l_2, l_3]^T$ can be calculated by the predicted state $X_{k+1|k}$ and the transformation $^I_C T$ using the projection Equation (6). The process is shown in Figure 4.

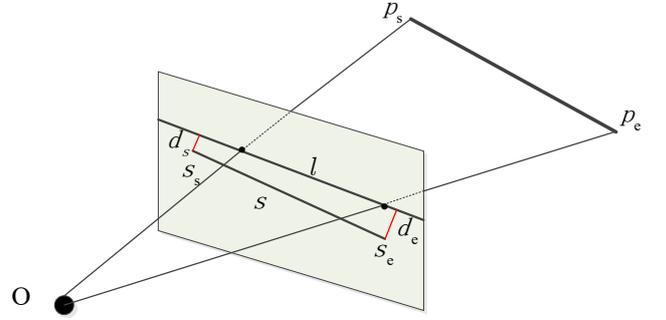

**FIGURE 4.** Projection and error of line segment

The re-projection error of lines is used for the update step to correct the predicted state. Here we define the measurement function containing the signed orthogonal distances $[d_s, d_e]^T$ from the detected endpoints $(s_s, s_e)$ to the predicted line $l$, which is the same as in [31]. The measurement function is shown

$$z = \begin{bmatrix} d_s \\ d_e \end{bmatrix} = \begin{bmatrix} l^T \cdot \overline{s}_s / \sqrt{l_1^2 + l_2^2} \\ l^T \cdot \overline{s}_e / \sqrt{l_1^2 + l_2^2} \end{bmatrix}, \quad (26)$$

where $\overline{s}_s$ and $\overline{s}_e$ are homogeneous coordination of $(s_s, s_e)$. We name this full observation function as $h(\cdot)$, which includes the transformation of coordination, line projection and calculation of error, so $h(\cdot)$ can be expressed by

$$z = h(X, s, Q_s) = h(T_c, \mathcal{L}, \mathcal{K}, s, Q_s), \quad (27)$$

where $Q_s$ is the covariance of measurement noise $n_s$.

(2) Construct cubature points and estimate the measurement

At time $k+1$, the measurement error of $m$ line features is calculated according to Equation (26). We use the cubature transformation to compute this. Same as the prediction step, the first step is to augment the state $X_{k+1|k}$ and square-root factor $S_{k+1|k}$ with the measurement noise $n_s$. The generator points $\{C^j\}, j = 1 \ldots 2l_m$, can be got based on cubature rule[26]. The number of $l_m$ is $8m + 15$, and the structure of $C^j$ is $[\xi^j_{k+1|k}, \tilde{b}^j_{k+1|k}, n^j_s]^T$. The calculation process for cubature points is shown by

$$\bar{X}^{k+1} \oplus C^j \triangleq \begin{cases} \exp(\xi^j_{k+1|k})\bar{\chi}_{k+1|k} \\ \bar{b}_{k+1|k} + \tilde{b}^j_{k+1|k} \\ n^j_s \end{cases}, j = 1 \ldots 2l_m. \quad (28)$$

The cubature points are transformed by measurement model using Equation(27), and the measurement cubature points can be got following

$$z^j_{k+1|k} = h(\exp(\xi^j_{k+1|k})\bar{\chi}_{k+1|k}, \mathcal{K}, s, Q_s) \quad (29)$$

Then based on the cubature transformation, we can got the mean of measurement estimation $\bar{z}_{k+1|k}$ and the error $e^j$ of cubature point, which is shown by

$$\bar{z}_{k+1|k} = \frac{1}{2l_m}\sum_{j=1}^{2l_m} z^j_{k+1|k}, \quad (30)$$

$$e^j = \frac{1}{\sqrt{2l_m}}(z^j_{k+1|k} - \bar{z}_{k+1|k}), j = 1 \ldots 2l_m. \quad (31)$$

The square-root factor of measurement error covariance $S_z$ is calculated by utilizing QR decomposition on the augmented error matrix $[e^1 \ldots e^{2l_m}; 0 \quad S_s]$, given by

$$[q, r] = qr([e^1 \ldots e^{2l_m}; 0 \quad S_s]), S_z = r^T, \quad (32)$$

where $r$ is the upper triangular matrix.

(3) Update the state and covariance

The updated state is performed using standard Kalman filtering algorithm. The Kalman gain is $K = P_{Xz}(S_z^T)^{-1}S_z^{-1}$, where $P_{Xz}$ is the cross-covariance matrix between the state and the measurement. $P_{Xz}$ can be calculated as

$$P_{Xz} = \frac{1}{2l_m}\sum_{j=1}^{2l_m}[\xi^j \quad \tilde{b}^j]^T(z^j - \bar{z}). \quad (33)$$

Different from the point-based VIO, the definition of measurement function (26) involves line transformation and projection and the innovation measurements. So $\bar{z}_{k+1|k}$ contains the innovation vector $[d_s, d_e]^T$ revealing the error between predicted line and measurement line, which is used to correct the state mean and covariance. Based on the Kalman filter, the innovation term $[\delta\xi, \delta b]^T$ is calculated by:

$$\begin{bmatrix} \delta\xi \\ \delta b \end{bmatrix} = K(0 - \bar{z}_{k+1|k}) \quad (34)$$

$\delta\xi$ is the innovation term for uncertainty $\xi$ of matrix Lie group $\chi = \exp(\xi)\bar{\chi}$. The approximation of the posterior $\chi$ can be expressed as $\chi^+ \approx \exp(\xi + \delta\xi)\bar{\chi} = \exp(\delta\xi)\chi$. $\delta b$ is the innovation term for IMU bias $b$. The corrected state can be calculated following

$$X_{k+1|k+1} = \begin{bmatrix} \chi_{k+1|k+1} \\ b_{k+1|k+1} \end{bmatrix} = \begin{cases} \exp(\delta\xi)\bar{\chi}_{k+1|k} \\ \bar{b}_{k+1|k} + \delta b \end{cases}. \quad (35)$$

The updated square-root factor of state error covariance can be calculated by performing QR decomposition on the error matrix, which is calculated as

$$[q, r] = qr\left(\begin{bmatrix} \xi^j_{k+1|k} \\ \tilde{b}^j_{k+1|k} \end{bmatrix} - Ke^j; 0 \quad S_s \right), \quad (36)$$

where $r$ is the upper triangular matrix. Then the updated state error covariance can be calculated following

$$P_{k+1|k+1} = S_{k+1|k+1}S^T_{k+1|k+1}, \text{ where } S_{k+1|k+1} = r^T. \quad (37)$$

## V. EXPERIMENTAL RESULT

We evaluate the proposed algorithm using the EuRoc MAV datasets [38], which are collected by an MAV in two different rooms and a large machine hall. The datasets contain stereo images from a global shutter camera at 20FPS and synchronized IMU measurements at 200Hz. We only use the images from the left camera. The ground truth trajectory and the calibration files are also provided. All of our experiments were performed on a laptop with Intel(R) core i7-5500CPU @2.40GHz 2.39GHz, 8GB RAM, and Matlab 2017A. In this section, we mainly evaluate the proposed VIO performance in the weak illumination scene and compare it with some state-of-the-art algorithms.

### A. INITIALIZATION OF 3D LINE

In the proposed algorithm, monocular images are used to update the filter, which cannot provide the scale information of 3D line. So we need to initialize the 3D line for the state matrix Lie group. The initial body pose corresponds to the ground truth values, and the spatial lines are reconstructed from two views following [39], which is also described in section III.(C). Figure 5 shows the results of initialization for the MH_05_difficult sequence. As shown in Figure 5(a), the lines in image are detected and matched using LSD [36] and LBD [37] algorithm, and the initialized spatial lines and the corresponding line in image are shown in Figure 5 (c) and (b). The corresponding parameters of the camera capturing this two matched images are shown in Table I.

During the experiment, a maximum 20 line features were maintained in the state of filter, and other tracked lines were saved in the candidate sequence of line management.

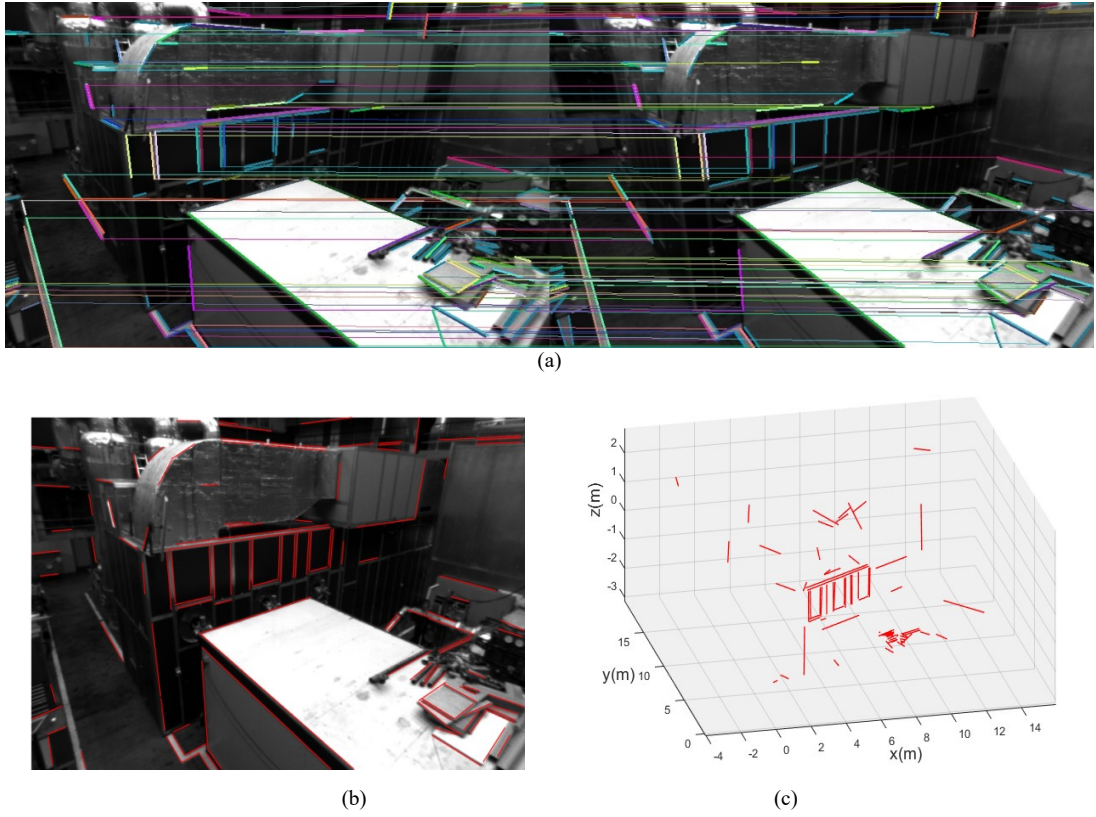

**Figure 5.** Matching and initialization of spatial lines. (a) The matched lines between two frames using LSD and LBD methods. (b) Extracting the lines in image for 3D estimation. (c) The results of spatial lines estimated in initialization.

TABLE I
CAMERA POSE AND INTRINSIC PARAMETERS IN INITIALIZATION

| Parameter | Data |
|---|---|
| Pose of left image | [ 0.9901, -0.1382, 0.0265; -0.0240, -0.3514, -0.9359; 0.1386, 0.9260, -0.3512 ] |
| Position of left image | [ 4.6091; 0.3022; 1.2372] |
| Pose of right image | [ 0.9904, -0.1359, 0.0249; -0.0251, -0.3542, -0.9348; 0.1359, 0.9252, -0.3542 ] |
| Position of right image | [4.5343; 0.4017; 1.2587] |
| Camera intrinsic matrix | [ 458.6540, 0, 0; 0, 457.2960, 0; 367.2150, 248.3750, 1.0000 ] |

### B. COMPARISION BETWEEN POINT-BASED AND LIEN-BASED ALGORITHMS

To demonstrate the advantage of the proposed algorithm in weak illumination scene and man-made environment, we compare L-SCKF-LG with the point-based algorithm RI-EKF[6] and SCKF-LG proposed in our recent work[40]. SCKF-LG algorithm has the same structure of filter, and the main difference is the landmarks in the state are points. The evaluation result can clearly show the different performance between line features and point features in illumination changing and texture-less scenes. The comparison with RI-EKF is mainly to prove the improvement of accuracy for using cubature points methods instead of calculating Jacobin matrix on Lie group.

These three algorithms are tested on MH_05_difficult sequence which contains challengeable scenes with illumination changing and low textured situations. Line-based VIO algorithm can provide rich geometry structure information with respect to the environment. A part of reconstructed 3D map of MH_05 is shown in Figure 6, and the structure information, such as wall and windows, is distinguished. This is useful for robot navigation.

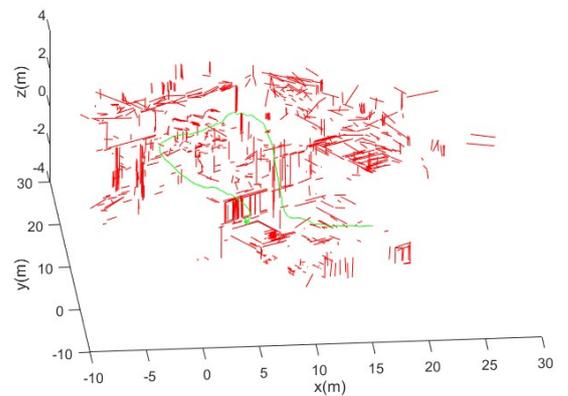

**FIGURE 6.** Part of estimated trajectory and map building by the proposed algorithm

The trajectories of L-SCKF-LG and SCKF-LG algorithms in the MH_05_difficult are shown in Figure 7. The black dash line represents the ground truth trajectory, and the blue dash line and red line represent the trajectories respectively for L-SCKF-LG and SCKF-LG algorithm. From the figure, we can see that two algorithms have similar performance in the beginning stage with fine illumination, and later the trajectory of proposed algorithm is closer to the ground truth when the MAV flew into the dark scene. For quantitative evaluation, we compare these three algorithms with loop closure proposed in [41]. This method adopts DIRD as the image descriptor, which is illumination robust and avoiding the influence of different features, so it is fair for the comparing algorithms. The RMSE(Root Mean Square Error) on position and attitude of robot is used as the evaluation criterion. The results of these algorithms running over 20 independence tests are shown in Figure 8 and Figure 9.

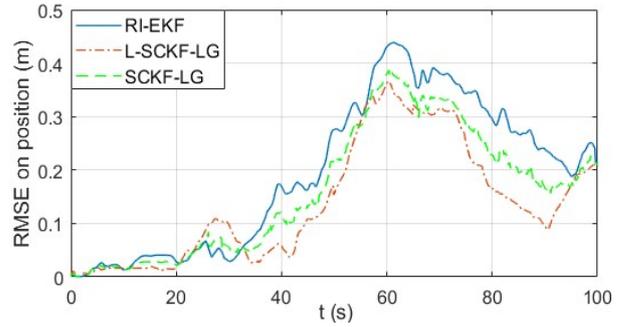

**FIGURE 8.** Comparison of Root Mean Square Error on position over time

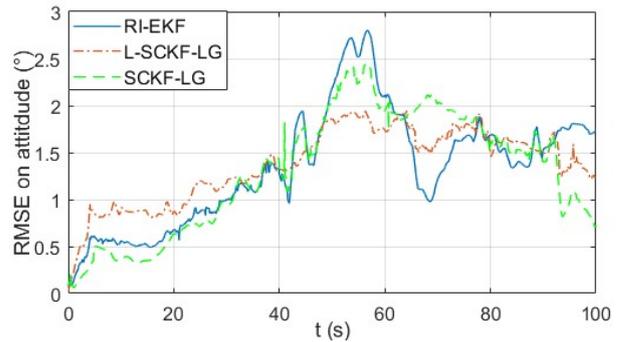

**FIGURE 9.** Comparison of Root Mean Square Error on attitude over time

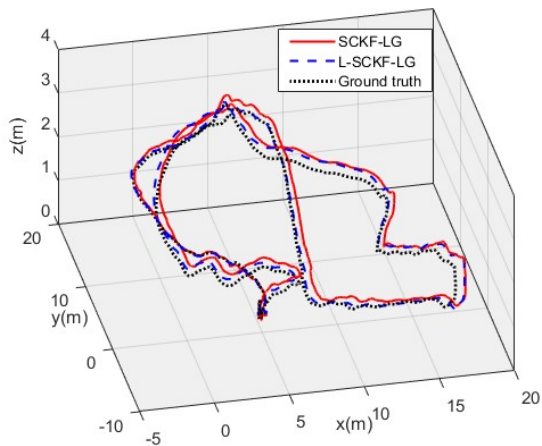

**FIGURE 7. The trajectories estimated by SCKF-LG and L-SCKF-LG algorithms**

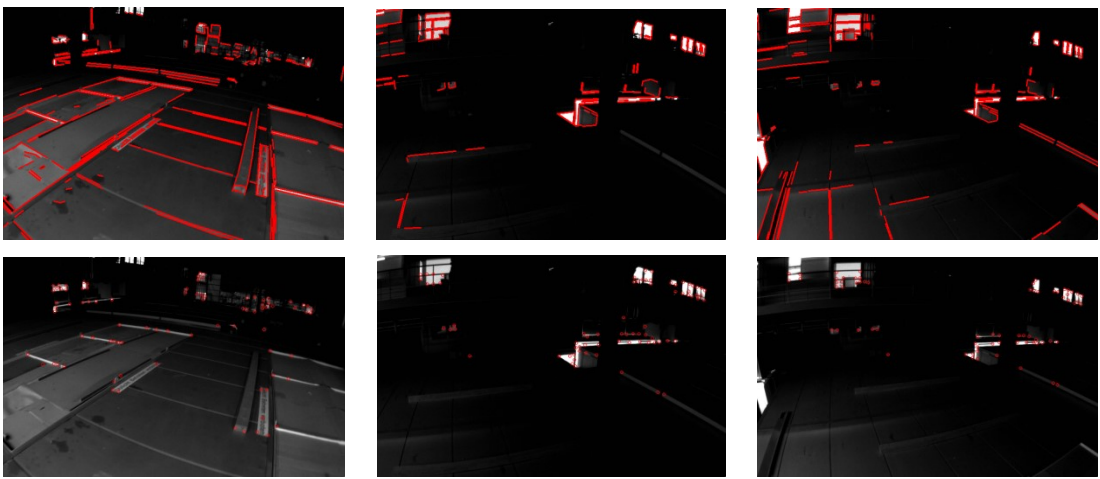

**FIGURE 10.** Comparison between points and line features extracted in dark environment

From Figure 8, we can see that position errors accumulate over time and gradually reduce after loop closure. The proposed algorithm doesn't perform better than SCKF-LG algorithm and RI-EKF algorithm in the first 30 seconds, but it leads to less average positon RMSE than other two algorithms in the remaining time. It is logical sine the MAV flies in the bright and textured scene in the beginning stage, and later flies into weak illumination and texture-less scene.

The detected point and line features in the images are shown respectively in figure 10. We can find that point features are rare and assemble in limited area in weak illumination scene, and line features are abundant and widely distributed. These advantages make line-based algorithms obtain better position performance in weak illumination and texture-less scene.

Figure 9 shows the attitude error comparison within three algorithms. From this figure, we can see that SCKF- LG algorithm performs better in the first 30 seconds, and later the attitude errors of RI-EKF and SCKF- LG increase rapidly and exceed the proposed algorithm in the middle stage. These attitude errors shrink after the loop closure. The variation trend of the proposed algorithm in attitude error is relatively stable, not as violent as the comparison algorithms. The result is logical since line projection is only related to the normal vector of 3D line, which is defined in formula (7). This means that the projection constraint is not sufficient from the normal vector of spatial line when the camera move up and down along the line. In the beginning, the MAV launches and lands vertically many times in the platform, which causes a large initial error. But the error grown trend of the proposed algorithm is lower than the compared algorithms, which proves that line-based algorithm is more robust in the weak illumination scene.

From these results, we may roughly conclude that point-based and line-based algorithms have similar performance in fine-illumination and textured environment, but line information can help more in weak-illumination and texture-less scene, which make line-based filter preform more robust and accurately. This conclusion needs to be further tested. In the next section, we will conduct experiments in more scenes and compare it with more algorithms.

*C. Comparison with other VIO algorithms on Euroc dataset*

Figure 11 shows the heat map of trajectories estimated by L-SCKF-LG and SCKF-LG running in V1_01_easy、V2_02_medium and MH_03_ medium sequences. The color bar shows the degree of position error. Comparing the three trajectories, we can see that the proposed algorithm does not outperform much in V1_01_easy and V2_02_medium, in which the scene contains abundant point and line features and fine illumination. But in MH_03_ medium sequence, the proposed algorithm perform better since the lighting condition changes greatly in this sequence and line features have more advantage in that situation.

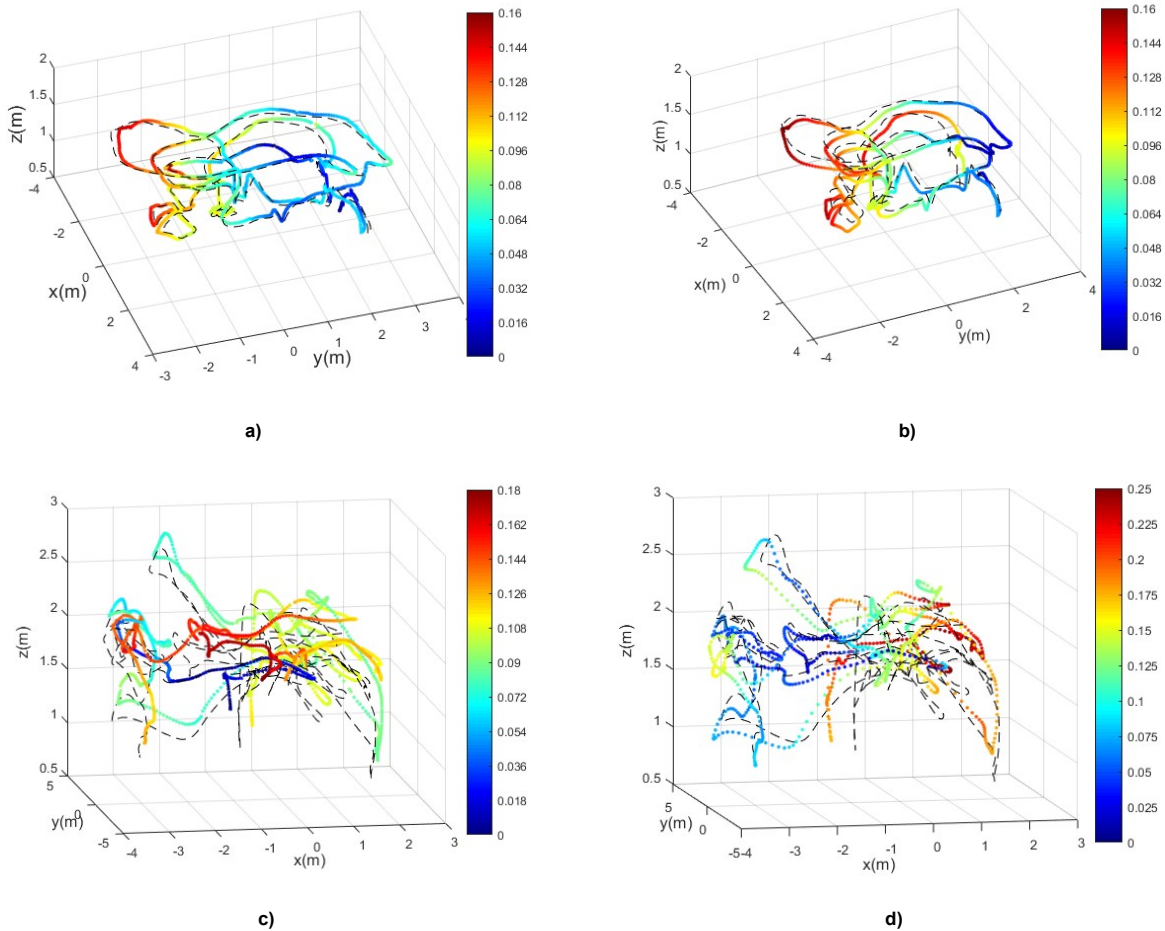

a)

b)

c)

d)

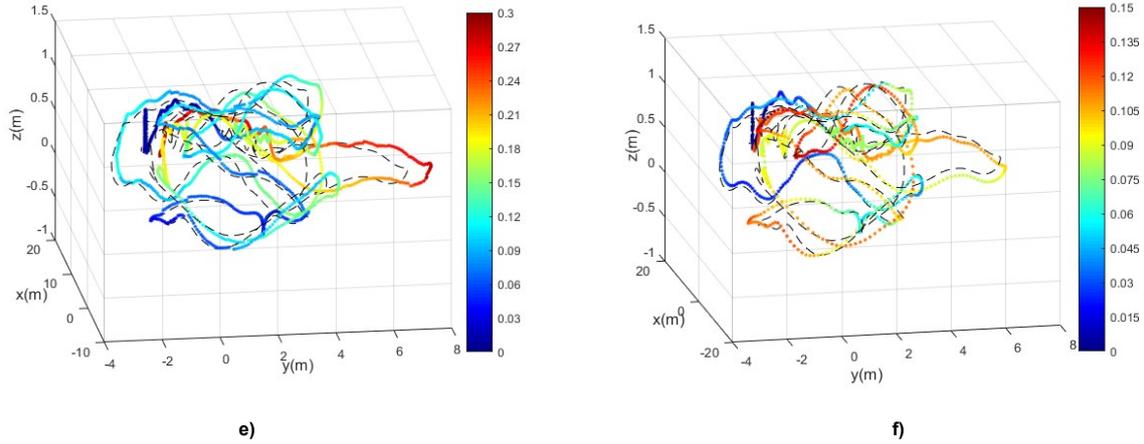

e)                                                                f)

FIGURE 11 Comparison of trajectory estimated between the proposed algorithm and SCKF-LG in different sequences.(a), (c),(e) are the trajectories respectively estimated by SCKF-LG running on V1_01, V2_02 and MH_03; (b), (d),(f) are the trajectories respectively estimated by the proposed algorithm running on V1_01, V2_02 and MH_03.

For evaluating the performance of the proposed algorithm, we compare it with other filter-based VIO algorithms on different Euroc sequences, including MSCKF, ROVIO, RI-EKF and SCKF-LG. MSCKF is a classic tightly-coupled filter-based VIO, the state vector of which contains several camera poses instead of 3D point positions. ROVIO is monocular VIO algorithm which use pixel intensity error of image patches to construct the observation equation in EKF. The results are shown in Table II and their histograms are shown in Figure 12. From these figures, we can see that the accuracy of ROVIO is the lowest in all sequences. And the proposed L-SCKF-LG algorithm achieves the best results in MH_03 and MH_05, where the scenes contain great illumination change and the MAV all flies into dark area. In fine-illumination scene, V101 and V202, the proposed algorithm has similar performance with others.

TABLE II RMSE ON POSITON ESTIMATED BY DIFFERENT ALGORITHMS RUNNING ON DIFFERENT EUROC SEQUENCES

| sequence | SCKF-LG | ROVIO | MSCKF | RI-EKF | L-SCKF-LG |
|---|---|---|---|---|---|
| V1_01_easy | 0.1325 | 0.2357 | 0.1236 | 0.1311 | 0.1324 |
| V2_02_medium | 0.1687 | 0.2832 | 0.2058 | 0.1825 | 0.2278 |
| MH_03_medium | 0.2532 | 0.4217 | 0.1674 | 0.3012 | 0.1489 |
| MH_05difficult | 0.3367 | 0.9538 | 0.4138 | 0.3500 | 0.3166 |

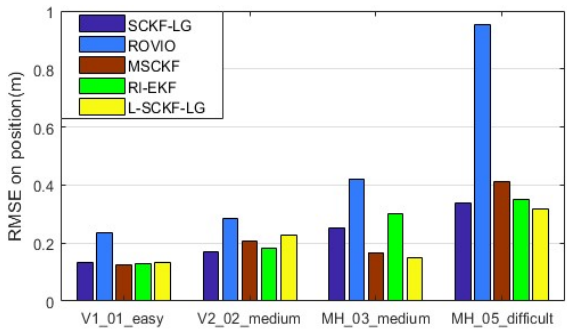

FIGURE 12. Comparison of position error estimated by different algorithms

From the results, we can say that the proposed algorithm works correctly and can effectively improve the robustness and accuracy of VIO in illumination-changing and texture-less environments. The line features included in state matrix are helpful in attitude estimation in particular scenes. However, we do acknowledge that the point features are also essential and perform well in good illumination and texture-rich scenes. So in order to make this algorithm applicable in wide conditions, we would have to fuse all these information in next step.

## VI. CONCLUSION

In this paper, we proposed a novel tightly-coupled monocular visual-inertial odometry algorithm upon the theory of invariant filtering and Cubature Kalman Filter. The system state is built by a high dimensional matrix Lie group consisting of body pose, velocity, and the line features of environments. The corresponding uncertainty is expressed by Lie algebra. In this method, the concept of covariance and cubature points are extended to the manifold space, and the prediction and measurement update are implemented on manifold. The hybridation of the standard SR-CKF with the invariant observer theory not only inherits the advantages of invariant filtering in consistency, but also avoids the complex Jacobian derivation process of high-dimensional matrix, which effectively improves the accuracy for VIO estimation. The re-projection errors of lines are built as observation function for updating the state , which improves the robustness for pose estimation in low-textured or illumination-changing environments. We compare the proposed algorithm with three state-of-the-art filter-based VIO methods on Euroc datasets, and the evaluation results verify the improvement of robustness and accuracy in the complex environment.

This method shows an expendable structure of state variable using high dimensional matrix Lie group, so users can readily insert additional measurements or parameters to estimate and avoiding complex computation of Jacobians. In order to extend its application in more complex situations, we plan to combine point and line features, and to add other measurement such as GPS into the state matrix in the future.